\documentclass[a4paper,sort&compress, fleqn]{cas-sc}

\usepackage[numbers]{natbib}
\usepackage{textcomp}
\usepackage{url}
\usepackage{verbatim}
\usepackage{graphicx}
\usepackage{amssymb}
\usepackage{amsmath}

\usepackage{amsthm}
\usepackage{array,booktabs}
\usepackage{multirow}
\usepackage{bm}
\usepackage{bbding}
\usepackage{pifont}
\usepackage{makecell}
\usepackage{xcolor}
\usepackage{array}
\usepackage{tabularx}
\usepackage{newfloat}
\usepackage{listings}
\usepackage{hyperref}
\usepackage[linesnumbered, ruled]{algorithm2e}
\usepackage{subcaption}
\usepackage{multicol}
\def\tsc#1{\csdef{#1}{\textsc{\lowercase{#1}}\xspace}}
\tsc{WGM}
\tsc{QE}
\tsc{EP}
\tsc{PMS}
\tsc{BEC}
\tsc{DE}
\begin{document}
\let\WriteBookmarks\relax
\def\floatpagepagefraction{1}
\def\textpagefraction{.001}
\let\printorcid\relax
\shorttitle{Information Fusion}

\shortauthors{Wang et~al.}

\title [mode = title]{DEMO: A Dynamics-Enhanced Learning Model for Multi-Horizon Trajectory Prediction in Autonomous Vehicles}                      
% \tnotemark[1,2]

\author[1,2]{Chengyue Wang}
\fnmark[1]
\credit{Conceptualization, Methodology, Experiment, Writing}

\author[1,3]{Haicheng Liao}
\fnmark[1]
\credit{Conceptualization, Methodology, Experiment, Writing}

\author[1,3]{Kaiqun Zhu}
\credit{Methodology, Review}

\author[4]{Guohui Zhang}
\credit{Methodology, Review}

\author[1,2,3]{Zhenning Li}
\cormark[1]
\ead{zhenningli@um.edu.mo}
\credit{Conceptualization, Methodology, Writing}

\affiliation[1]{organization={State Key Laboratory of Internet of Things for Smart City, University of Macau}, city={Macau SAR}, country={China}}

\affiliation[2]{organization={Department of Civil and Environmental Engineering, University of Macau}, city={Macau SAR}, country={China}}

\affiliation[3]{organization={Department of Computer and Information Science, University of Macau}, city={Macau SAR}, country={China}}

\affiliation[4]{organization={ Department of Civil, Environmental and Construction Engineering, University of Hawaii at Manoa}, city={Honolulu, Hawaii}, country={U.S.}}

\cortext[cor1]{Corresponding author; $^{1}$Equally Contributed}


\begin{abstract}
Autonomous vehicles (AVs) rely on accurate trajectory prediction of surrounding vehicles to ensure the safety of both passengers and other road users. Trajectory prediction spans both short-term and long-term horizons, each requiring distinct considerations: short-term predictions rely on accurately capturing the vehicle's dynamics, while long-term predictions rely on accurately modeling the interaction patterns within the environment. 
However current approaches, either physics-based or learning-based models, always ignore these distinct considerations, making them struggle to find the optimal prediction for both short-term and long-term horizon.
In this paper, we introduce the \textbf{D}ynamics-\textbf{E}nhanced Learning \textbf{M}\textbf{O}del (\textbf{DEMO}), a novel approach that combines a physics-based Vehicle Dynamics Model with advanced deep learning algorithms. DEMO employs a two-stage architecture, featuring a Dynamics Learning Stage and an Interaction Learning Stage, where the former stage focuses on capturing vehicle motion dynamics and the latter focuses on modeling interaction. By capitalizing on the respective strengths of both methods, DEMO facilitates multi-horizon predictions for future trajectories. 
Experimental results on the Next Generation Simulation (NGSIM), Macau Connected Autonomous Driving (MoCAD), Highway Drone (HighD), and nuScenes datasets demonstrate that DEMO outperforms state-of-the-art (SOTA) baselines in both short-term and long-term prediction horizons.

% \noindent\texttt{\textbackslash begin{abstract}} \dots 
% \texttt{\textbackslash end{abstract}} and
% \verb+\begin{keyword}+ \verb+...+ \verb+\end{keyword}+ 
% which
% contain the abstract and keywords respectively. 

% \noindent Each keyword shall be separated by a \verb+\sep+ command.
\end{abstract}

\begin{keywords}
Autonomous Driving \sep Trajectory Prediction  \sep Dynamics-based Model \sep Learning-based Model \sep Data Fusion
\end{keywords}

\maketitle
\begin{sloppypar}

\section{Introduction}
Autonomous vehicles (AVs) are poised to revolutionize transportation by enhancing safety, efficiency, and reliability \cite{lu2025hyper}. A critical capability for safe operation in dynamic traffic environments is the accurate prediction of future trajectories of surrounding agents—including vehicles, pedestrians, and cyclists \cite{lan2024hi}. Such accurate trajectory prediction allows AVs to anticipate hazards, adapt to changing traffic conditions, and plan safe, real-time navigation strategies \cite{li2024context}. However, the diverse temporal scales of driving scenarios necessitate dividing trajectory prediction into two key horizons: \textbf{short-term} ($\leq2$ seconds) and \textbf{long-term} ($>$2 seconds), each presenting unique technical challenges and requiring different modeling approaches \cite{chen2022intention, chandra2020forecasting}.

In the short-term horizon, AVs must handle immediate, high-stakes changes such as sudden braking, abrupt lane changes, or unexpected obstacle appearances. Errors within this brief window can quickly compound, making rapid and precise predictions critical for maintaining safety and stability. Physics-based models are particularly well-suited for short-term predictions because they capture instantaneous relationships among variables such as force, acceleration, and velocity. This direct modeling provides a structured, deterministic framework that enables AVs to respond rapidly to immediate changes without the computational overhead associated with more complex models \cite{anderson2021kinematic, xie2017vehicle}. During high-speed or emergency maneuvers, physics-based models offer predictable and stable predictions based on well-established dynamic principles, ensuring consistent performance in safety-critical situations.

Conversely, in the long-term horizon, AVs engage in strategic planning where they navigate broader interactions and scenarios such as merging into traffic, anticipating pedestrian crossings, or adjusting to evolving traffic flow patterns. Learning-based models are adept at capturing and generalizing from extensive historical data, making them well-suited for modeling complex, multi-agent interactions in the long-term horizon \cite{mo2024heterogeneous, ren2024emsin}. These models allow for initial predictions that can be refined as new data becomes available, supporting ongoing adjustment and optimization of long-term planning. The adaptability of learning-based models enables AVs to understand trends, intentions, and social dynamics over time, which is essential for strategic decision-making that balances immediate reactivity with long-term objectives \cite{feng2023macformer, geng2023adaptive}.

Despite advancements in both physics-based and learning-based models, most existing trajectory prediction approaches are confined to one paradigm, each optimized for a single horizon. Physics-based models excel in short-term responses but struggle to generalize in complex, multi-agent environments. Conversely, learning-based models effectively capture long-term interactions but often lack the precision required for real-time control in short-term scenarios \cite{huang2022survey, houenou2013vehicle}. While some studies have attempted to integrate these approaches, they often fail to fully exploit the complementary strengths of both methods. This dichotomy underscores the need for an integrative model that seamlessly blends the immediate responsiveness of physics-based dynamics with the adaptive capabilities of learning-based models, thereby addressing both short-term precision and long-term adaptability.

\begin{figure}[tbp]
  \centering
\includegraphics[width=0.9\linewidth]{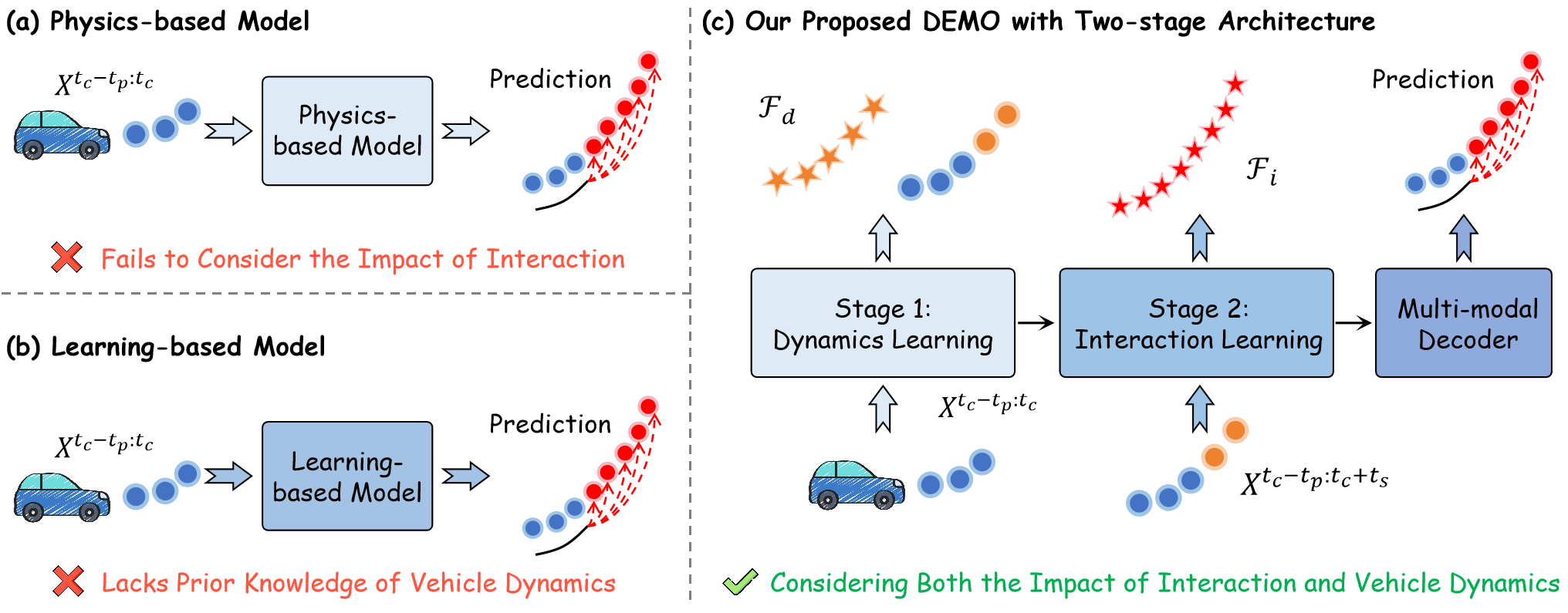} 
  \caption{Comparison between the traditional Physics-based model (a), Learning-based model (b), and our proposed DEMO (c). Unlike previous approaches, DEMO employs a two-stage architecture for hierarchical prediction, integrating a Dynamics Learning Stage with an Interaction Learning Stage.}
  \label{toutu} 
\end{figure}

To address this gap, we propose \textbf{DEMO}, a \textbf{D}ynamics-\textbf{E}nhanced Learning \textbf{M}\textbf{O}del that synergistically combines the strengths of physics-based and learning-based approaches. As shown in Fig. \ref{toutu}, DEMO features a two-stage architecture: the \textbf{Dynamics Learning Stage} and the \textbf{Interaction Learning Stage}. The Dynamics Learning Stage integrates a physics-based Dynamic Bicycle Model with a Dynamic Conditional Variational Autoencoder (DynCVAE) to precisely capture vehicle dynamics within the short-term horizon, delivering the high-fidelity, low-latency responses essential for short-term control. The Interaction Learning Stage employs deep learning methods with cross-modal fusion and spatio-temporal encoding to model the complex, long-term interactions characteristic of the long-term horizon. This design enables DEMO to perform precise, immediate short-term maneuvers while effectively managing long-term, socially interactive scenarios, thus fulfilling the diverse requirements of AV trajectory prediction.

The primary contributions of this paper are as follows:

\begin{itemize}
    \item We design a Dynamics Learning Stage that integrates a Dynamic Bicycle Model with a Dynamic Conditional Variational Autoencoder (DynCVAE). This synergistic fusion of vehicle dynamics modeling with deep learning techniques enables DEMO to capture intricate vehicle motion dynamics, ensuring high accuracy in short-term predictions essential for immediate AV response.
    
    \item To support accurate long-term predictions, we develop an Interaction Learning Stage that incorporates cross-modal fusion and spatial-temporal encoder. This stage enables DEMO to model complex environmental and social interactions, allowing the model to adapt to diverse, multi-agent scenarios and facilitate strategic decision-making in the long-term horizon.
    
    \item Evaluated on multiple well-established datasets, including NGSIM, HighD, MoCAD, and nuScenes, DEMO significantly outperforms state-of-the-art (SOTA) baselines across diverse driving scenarios. Evaluations conducted on a limited 50\% subset of the dataset, along with inference speed comparison, further highlight DEMO's efficiency and suitability for real-world applications.
    
\end{itemize}

The structure of this paper is as follows: In Section \ref{Related work}, we review the relevant literature in the field. Section \ref{Methodology} provides a detailed explanation of DEMO's methodology. Section \ref{Experiment} presents a comprehensive evaluation of DEMO's performance. Finally, Section \ref{Conclusion} concludes this study with a summary.

\section{Related Work}\label{Related work}
In physics-based trajectory prediction, vehicle kinematic and dynamic models serve as mathematical frameworks for describing the evolution of a vehicle's motion state over time. Beyond trajectory prediction, vehicle dynamics and kinematic models are widely applied in various fields, including autonomous driving \cite{marzbani2019autonomous}, vehicle queuing \cite{liu2021joint}, and vehicle control \cite{zhang2023new}. Initially, the simplicity of kinematic models contributed to their widespread application. Over time, however, researchers recognized the limitations of kinematic models, leading to the development of numerous dynamic models tailored to diverse traffic scenarios \cite{yang2013overview, liu2021VTC}. For instance, Liu et al. \cite{liu2020lateral} proposed a two-stage vehicle dynamic model specifically for lateral control. This model first estimates the trajectory’s radius of curvature using the least squares method and then applies fixed-structure feedback control to improve motion control. Ge et al. \cite{ge2021numerically} developed a discrete vehicle dynamic model for intelligent vehicles based on the backward Euler equation. Experimental results demonstrated that this model not only exhibits numerical stability but also significantly enhances prediction accuracy compared to kinematic models. Bellegarda et al. \cite{bellegarda2021dynamic} adopted a different approach by integrating vehicle dynamics with a kinematic model within a model predictive control framework, focusing on drift control. In summary, vehicle dynamic models consider the complex mechanisms underlying subtle parameter changes in vehicle motion, enabling them to discern intricate dynamic patterns. This capability is crucial for accurately capturing vehicle motion patterns. However, as these models focus solely on the vehicle’s intrinsic attributes, they overlook interactions among traffic participants within the environment \cite{huang2022survey}, limiting their effectiveness in accurately predicting long-term trajectories.

In learning-based trajectory prediction, accurately modeling interactions within the environment is crucial for enhancing predictive accuracy. Initially, researchers developed numerous trajectory prediction models based on classical machine learning, achieving significant improvements in accuracy compared to physics-based methods. However, the diversity of traffic scenarios and the inherent uncertainty of driving behaviors introduce substantial complexity to interaction modeling \cite{liao2024cognitive2}, leading to stagnation in machine learning algorithms that rely heavily on handcrafted feature engineering. To address this complexity, researchers have introduced various deep learning-based trajectory prediction models \cite{park2024t4p, liao2024characterized}, leveraging the large parameter capacities of deep learning algorithms to model intricate interaction patterns. Transformers, originally designed for natural language processing, have become dominant in trajectory prediction due to their attention mechanisms, which capture dependencies between any two tokens within a sequence. Initially, researchers focused on vehicle-to-vehicle interactions: Chen et al. \cite{chen2022intention} used attention mechanisms to extract spatial interactions between vehicles and then applied attention modeling to capture temporal dependencies, yielding a comprehensive spatial-temporal interaction model. Subsequently, researchers gradually recognized that in addition to direct vehicle-to-vehicle interactions, interactions between vehicles and infrastructure also significantly influence vehicle behavior. Zhang et al. \cite{zhang2022trajectory} developed a trajectory prediction model based on a Graph Attention Transformer to model agent-agent and agent-infrastructure interactions through attention mechanisms. In addition to infrastructure, vehicle movement must also comply with road topology; Geng et al. \cite{geng2023adaptive} introduced a hierarchical Transformer network designed to capture multi-dimensional dependencies, including temporal, spatial, and vehicle-road topology dependencies. Unlike previous work implicitly embedded road topology, Feng et al. \cite{feng2023macformer} explicitly incorporated map constraints and developed a Map-Agent Coupling Transformer (MacFormer) for trajectory prediction. More recent approaches have explored integrating domain-specific priors within Transformer architectures, allowing for interpretable trajectory predictions. For instance, Geng et al. \cite{geng2023physics} proposed PIT-IDM, combining the Physics-Informed Transformer with the Intelligent Driver Model to predict longitudinal vehicle trajectories. Liao et al. \cite{liao2024human} introduced a human-inspired trajectory prediction model that effectively models spatial dependencies in driving scenarios through a Transformer framework. Recognizing the inherent differences between short-term and long-term trajectory prediction in human motion forecasting, Lin et al. \cite{lin2024progressive} proposed the PPT framework, achieving SOTA performance. As computational power advances, large language models (LLMs) based on transformer architectures have begun to emerge, demonstrating strong capabilities across multiple domains. Researchers are now exploring the use of LLMs to understand complex dependencies in driving scenarios. Lan et al. \cite{lan2024traj} proposed a trajectory prediction model based on pre-trained LLMs, leveraging their advanced scene understanding to significantly enhance trajectory prediction accuracy. Bae et al. \cite{bae2024can} transformed trajectory prediction, originally a numerical regression task, into a question-answering task, developing a language-based multimodal predictor, LMTraj.

\begin{figure}[tbp]
  \centering
\includegraphics[width=0.85\linewidth]{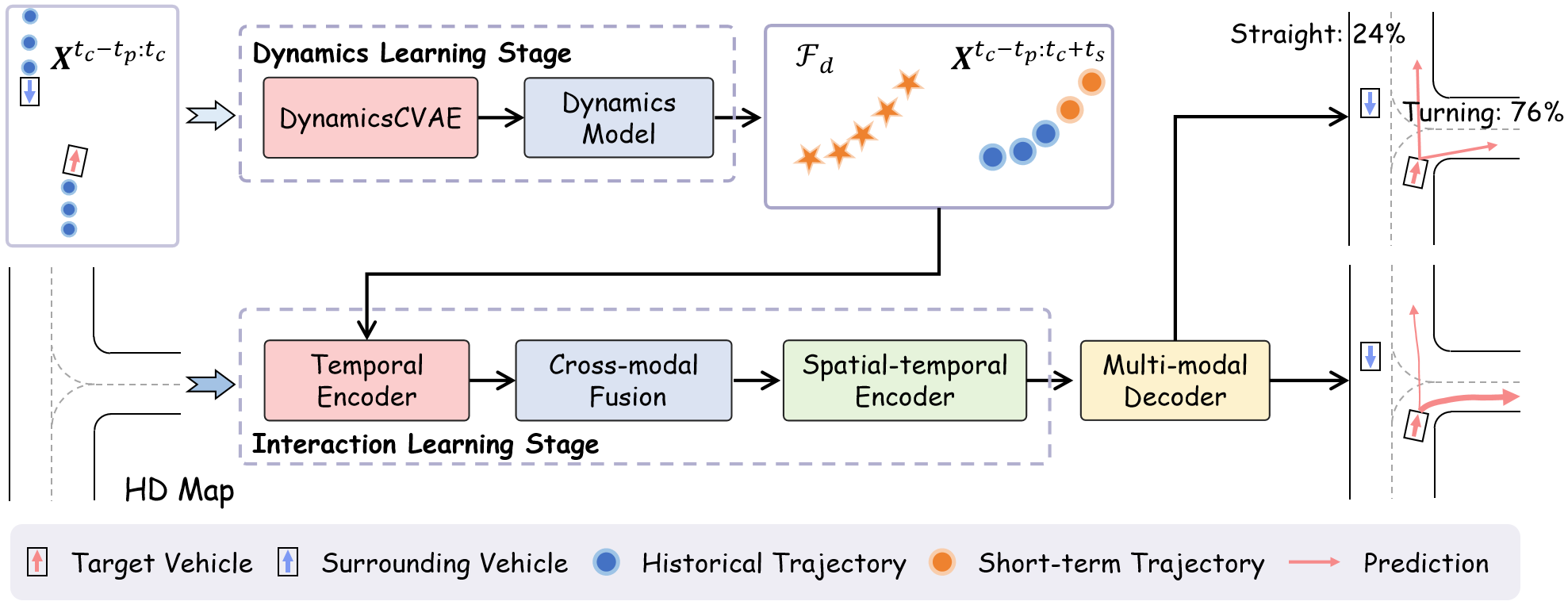} 
  \caption{Overall architecture of DEMO. 
  In the Dynamics Learning Stage, DEMO takes the historical trajectory as input, processes it through the DynCVAE and the Dynamic Bicycle Model, and ultimately generates the dynamic features and short-term trajectory. In the subsequent Dependencies Learning Stage, the model leverages the enhanced output from the previous stage, along with HD map data, to model the interaction present in the scene. Finally, a Multi-modal Decoder integrates the outputs from both stages to generate multi-modal predictions.}
  \label{framework} 
\end{figure}

\section{Methodology}\label{Methodology}
\subsection{Problem Formulation}\label{Problem Formulation}
The primary objective of this work is to develop a model that accurately predicts the future trajectories of vehicles surrounding an AV. In the DEMO, the vehicle to be predicted is referred to as the \textit{target vehicle}, while all other vehicles are collectively termed as \textit{surrounding vehicles}. The model's input consists of the HD map $\bm{M}$ of the scene, along with the historical trajectory of both the target vehicle $\bm{X}^{t_c-t_p:t_c}_0$ and the surrounding vehicles $\bm{X}^{t_c-t_p:t_c}_0$ over a past horizon $t_p$. The output of the model is the predicted future trajectory of the target vehicle $\bm{Y}^{t_c+1:t_c+t_f}_0$ over a specified prediction horizon $t_f$ starting from the current moment $t_c$.

\subsection{Overview of the Model Architecture}
Our proposed DEMO incorporates a novel two-stage architecture, as illustrated in Figure \ref{framework}, comprising the Dynamics Learning Stage and Interaction Learning Stage for trajectory predictions. The former stage is dedicated to generate the short-term dynamic feature $\mathcal{F}_{d}$ as well as short-term trajectory $\bm{X}^{t_c+1:t_c+t_s}_{0:n}$. This facilitates a refined understanding of vehicle dynamics patterns within a short-term horizon $t_s$. The latter instead, processes the historical trajectory $\bm{X}^{t_c-t_p:t_c}_{0:n}$ and short-term trajectory $\bm{X}^{t_c+1:t_c+t_s}_{0:n}$ to obtain the interaction feature $\mathcal{F}_{i}$ to modeling the interaction. Given the dynamic feature $\mathcal{F}_{d}$ and interaction feature $\mathcal{F}_{i}$, facilitating the multi-modal decoder to generate predictions for the entire prediction horizon.

\subsection{Dynamics Learning Stage}
The objective of the Dynamics Learning Stage is to learn the dynamics patterns within the short-term horizon. Vehicle motion dynamics are crucial for short-term prediction; therefore, we develop a hybrid approach that fuses a dynamics-based method, namely the Dynamic Bicycle Model, with a learning-based method, namely DynCVAE. This hybrid approach allows us to capture precise dynamic features. We begin this section with preliminaries of the Dynamic Bicycle Model to provide readers with the necessary background knowledge. It is followed by a description of how to generate the dynamic feature $\mathcal{F}_{d}$.
\subsubsection{Preliminaries}
The Dynamic Bicycle Model effectively represents vehicle dynamics, commonly used in autonomous driving for control system design and analysis \cite{yang2013overview}. By consolidating a four-wheel vehicle into an equivalent two-wheel system, it captures essential lateral and longitudinal dynamics with reduced complexity. Striking a balance between accuracy and computational efficiency, the model is more precise than simple kinematic models yet less complex than full 3D representations, making it suitable for real-time applications. It realistically simulates diverse driving scenarios, such as acceleration, deceleration, and cornering, crucial for short-term trajectory prediction. Moreover, the model’s flexibility allows extensions for factors like load transfer, aerodynamic forces, or individual wheel dynamics, enhancing its adaptability across various configurations and driving conditions \cite{ge2021numerically}. This balance of precision, realism, and flexibility makes the Dynamic Bicycle Model highly effective for trajectory prediction tasks.

Originally, the Dynamic Bicycle Model is formulated using ordinary differential equations. In this model, the variables are classified into three categories: attributes, state, and control variables. Attributes refer to inherent properties of the vehicle, including vehicle mass $m$, moment of inertia $I_z$, distance from the center of mass of the car to the front axles $l_f$ and rear axles $l_r$, and cornering stiffness of front $k_f$ and rear wheels $k_r$. State variables, which evolve over time, including the x-coordinate $x$ (forward direction) and y-coordinate $y$ of the vehicle position, the x-direction and y-direction components of the vehicle speed $v_x$ and $v_y$, yaw angle $\varphi$, yaw rate $\omega$. Control variables are external commands required to drive the model, including front wheel steering angle $\delta$, and acceleration command $a$. The system of differential equations that governs the Dynamic Bicycle Model is expressed as follows:
\begin{equation}\label{dyna1}
\begin{bmatrix}
\displaystyle \dot{x} \\
\displaystyle \dot{y} \\
\displaystyle \dot{v}_x \\
\displaystyle \dot{v}_y \\
\displaystyle \dot{\varphi} \\
\displaystyle \dot{\omega}
\end{bmatrix}
=
\begin{bmatrix}
\displaystyle v_x \cos \varphi-v_y \sin \varphi \\
\displaystyle v_x \sin \varphi+v_x \cos \varphi \\
\displaystyle a+v_x \omega-\frac{1}{m} k_f\left(\frac{v_y+l_f \omega}{v_x}-\delta\right) \sin \delta \\
\displaystyle -v_x \omega+\frac{1}{m}\left(k_f\left(\frac{v_y+l_f \omega}{v_x}-\delta\right) \cos \delta+k_r \frac{v_y-l_r \omega}{v_x}\right)\\
\displaystyle \omega \\
\displaystyle \frac{1}{I_z}\left(l_f k_f\left(\frac{v_y+l_f \omega}{v_x}-\delta\right) \cos \delta-l_r k_r \frac{v_y-l_r \omega}{v_x}\right)
\end{bmatrix}
\end{equation}

\subsubsection{Dynamic Feature Generation}
To effectively integrate the Dynamic Bicycle Model into the trajectory prediction task, we employ a discrete system approach, as outlined in \cite{ge2021numerically}. In this approach, we redefine the yaw angle $\varphi$ and yaw rate $\omega$ as control variables, departing from their traditional role as state variables in Equation \ref{dyna1}. This adjustment enables greater flexibility across different dataset configurations. Formally, the dynamic system is represented by:
\begin{equation}\label{dynamic_bicycle}
\begin{bmatrix}
\displaystyle x^{t+1} \\
\displaystyle y^{t+1} \\
\displaystyle v_{x}^{t+1} \\
\displaystyle v_{y}^{t+1} \\
\end{bmatrix}
=
\begin{bmatrix}
\displaystyle x^t + \Delta t (v_{x}^{t} \cos \varphi^t - v_{y}^{t}  \sin \varphi^t) \\
\displaystyle y^t + \Delta t (v_{y}^{t} \cos \varphi^t + v_{x}^{t} \sin \varphi^t) \\
\displaystyle v_{x}^{t} + \Delta t a^t \\
\displaystyle \frac{m v_{x}^{t} v_{y}^{t} + \Delta t \left( l_f k_f - l_r k_r \right) \omega^t - \Delta t k_f \delta^t v_{x}^{t} - \Delta t m {v_{x}^{t}}^2 \omega^t}
{m v_{x}^{t} - \Delta t (k_f + k_r)} \\
\end{bmatrix}
\end{equation}
where $\Delta t$ is the time interval between time step $t$ and step $t+1$.

Unlike traditional approaches that rely on dynamics models to directly predict future trajectories, our approach emphasizes that the control variables $\bm{C}^t=[\varphi^t, \omega^t, \delta^t, a^t]$, also provide the model with crucial insights into the vehicle's underlying dynamic behavior. Moreover, the control variables $\bm{C}^t$ and state variables (trajectory) $\bm{X}^t=[x^{t},y^{t},v_x^{t},v_y^{t}]$ can be transformed via the Equation \ref{dynamic_bicycle}. Therefore, the key to generating the dynamic features $\mathcal{F}_{d}$ lies in accurately estimating the control variables $\bm{C}$ of the short-term horizon. To this end, we designed a DynCVAE based on the vehicle dynamics model to estimate the control variable. Given the historical trajectory $\bm{X}$, this DynCVAE generates control variables $\bm{C}$ with the generation distribution $p(\bm{C}^t|\bm{X}^t,\bm{z}^t)$, where $\bm{z}$ represents latent control variables. It is evident that the control variables $\bm{C}^t=\Upsilon^{-1}(\bm{X}^{t},\bm{X}^{t+1})$, where $\Upsilon^{-1}$ denotes the inverse form of Equation \ref{dynamic_bicycle}. Therefore, we could reasonably hypothesize that the latent control variables could be accurately inferred with the posterior distribution $q(\bm{z}^t|\bm{X}^t, \bm{X}^{t+1})$. Since future states $\bm{X}^{t+1}$ are unavailable during inference, we introduce a Kullback–Leibler (KL) divergence loss \cite{xu2023context}, $\mathcal{L}_\textit{KL}$, to align the posterior distribution $q(\bm{z}^t | \bm{X}^t, \bm{X}^{t+1})$ with the prior distribution $p(\bm{z}^t | \bm{X}^t)$. This alignment allows the prior distribution to accurately model the latent control variables $\bm{z}$ during inference. Formally,
\begin{equation}
\mathcal{L}_\textit{KL}=\frac{1}{t_s} \sum_{t=t_c}^{t_c+t_s-1}\left[q\left(\mathbf{z}^t \mid \bm{X}^t, \bm{X}^{t+1}\right) \| p\left(\mathbf{z}^t \mid \bm{X}^t\right)\right]
\end{equation}

The iterative generation process begins by embedding the historical trajectory $\bm{X}^{t_c-t_p:t_c}$, producing $\mathring{\bm{X}} = \phi_{\textit{MLPs}}(\bm{X}^{t_c-t_p:t_c})$, where $\phi_{\textit{MLPs}}$ denotes the multilayer perceptions (MLPs). We initialize the system with $\bm{C}^{t_c} \sim p(\cdot | \mathring{\bm{X}}, \bm{z}^{t_c})$, where $\bm{z}^{t_c} \sim p(\cdot | \mathring{\bm{X}})$. For subsequent steps, we compute future states as $\bm{X}^{t+1} = \Upsilon(\bm{X}^t, \bm{C}^t)$ and update the latent control variables $\bm{z}^{t+1} \sim p(\cdot | \bm{X}^{t+1})$, followed by the control variables $\bm{C}^{t+1} \sim p(\cdot | \bm{X}^{t+1}, \bm{z}^{t+1})$. This iterative process allows for the generation of both short-term trajectory $\bm{X}^{t_c+1:t_c+t_s}_{0:n}$ and control variables $\bm{C}^{t_c:t_c+t_s-1}_{0:n}$. Finally, the short-term trajectory and control variables are projected into a high-dimensional space via MLPs, generating the dynamic features $\mathcal{F}_{d}$. Given the absence of ground truth for the control variables in the datasets, we design a dynamics-informed loss function $\mathcal{L}_\textit{DI}$ based on the Equation \ref{dynamic_bicycle}, to supervise the training of the DynCVAE.
Formally,
\begin{equation}
\mathcal{L}_\textit{DI}= \bm{X}^{t} - \bm{\hat{X}}^{t} = \bm{X}^{t} -\Upsilon(\bm{X}^{t-1},\bm{\hat{C}}^{t-1})
\end{equation}
where $\bm{\hat{C}}^{t-1}$ and $\bm{\hat{X}}^{t}$ denotes the generated control variables and trajectory.

\subsection{Interaction Learning Stage}
This stage is designed to accurately model the interaction pattern in the environment. It utilizes a deep learning-based architecture to respectively capture the dependencies between vehicles and their surrounding environment from spatial and temporal dimensions, enabling the model to discern spatial-temporal interactions in the environment. Initially, we input the historical trajectories $\bm{X}^{t_c-t_p:t_c}_{0:n}$ of the target vehicle and its surrounding vehicles, as well as the short-term trajectory $\bm{X}^{t_c+1:t_c+t_s}_{0:n}$ predicted by the first stage, into a temporal encoder. This encoder identifies temporal-attentive segments within the trajectories, producing encoded features for both the target $\mathcal{F}_{t}$ and surrounding vehicles $\mathcal{F}_{s}$. Next, we introduce an innovative cross-modal fusion that converts these features with HD map data $\bm{M}$ into a high-dimensional spatial space. In this space, they are further fused and integrated to generate cross-modal features $\mathcal{F}_{c}$ that guide the model to infer spatial interaction patterns. Finally, a spatio-temporal encoder captures the spatial-temporal interactions, producing interaction features, $\mathcal{F}_{i}$, that encompass rich spatial and temporal dependencies.

\subsubsection{Temporal Encoder}
The trajectory data of vehicles is essentially a time-series sequence, making dependencies modeling along the temporal dimension crucial. This is typically achieved with a transformer-based architecture. To address the computational inefficiency of the transformer framework on long sequences, we employ a highly efficient temporal encoder based on the state space model (SSM) \cite{aoki2013state} to capture temporal dependencies across various time steps and to recognize distinct motion patterns of different vehicles.

Specifically, the temporal encoder comprises a normalization layer, a Mamba block \cite{aoki2013state}, and a fully connected (FC) layer. Firstly, the input historical trajectory $\bm{X}^{t_c-t_p:t_c}_{0:n}$ is combined with the short-term trajectory $\bm{X}^{t_c+1:t_c+t_s}_{0:n}$ obtained in the first stage and then normalized. Subsequently, the merged input undergoes encoding in the Mamba block, which recursively scans the input trajectory sequences, continuously updating and capturing temporal dependencies throughout the sequence. Moreover, it incorporates a selection mechanism to determine which segments of the sequence contribute to the hidden state, allowing the model to concentrate on pivotal moments and dynamically process scene semantics with flexibility. This is followed by an FC layer and a sigmoid activation function, resulting in temporally-aware vehicle features $\mathcal{F}_{v}$. Formally,
\begin{equation}
    \mathcal{F}_{v} = [\mathcal{F}_{t},\mathcal{F}_{s}] = \phi_{\textit{sigmoid}}\left(\phi_{\textit{FC}}\left(\phi_{\textit{MB}}(\phi_{\textit {LN}}(\bm{X}^{t_c-t_p:t_c}_{0:n}\|\bm{X}^{t_c+1:t_c+t_s}_{0:n}))\right) \right) 
\end{equation}
where the vehicle features include the features for the target $\mathcal{F}_{t}$ and the surrounding vehicles $\mathcal{F}_{s}$. Moreover,  $\phi_{\textit {LN}}$, $\phi_{\textit{MB}}$, $\phi_{\textit{FC}}$, and $\phi_{\textit{sigmoid}}$ denote the normalization layer, Mamba block, FC layer, and sigmoid activation function, respectively.

\subsubsection{Cross-modal Fusion}
\begin{figure}[tbp]
  \centering
\includegraphics[width=0.7\linewidth]{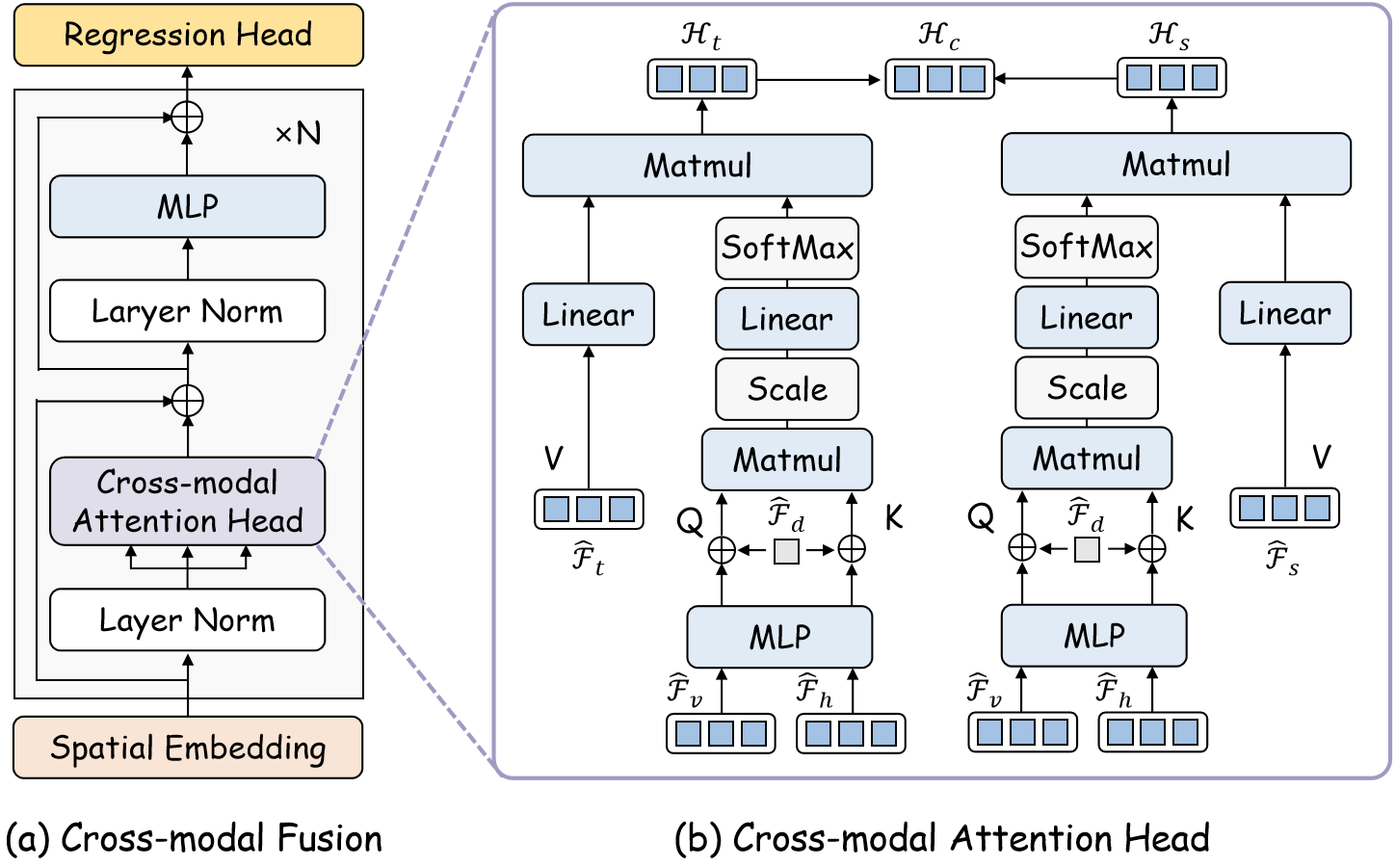} 
  \caption{Illustration of the Cross-modal Fusion. Panel (a) presents the pipeline of the Cross-modal Fusion process, while Panel (b) provides a detailed illustration of the mechanism of the Cross-modal Attention Head.}
  \label{cross} 
\end{figure}

Figure \ref{cross} (a) presents the comprehensive pipeline of the proposed cross-modal fusion framework. Taking vehicle features, HD map data, and dynamic features as input, the core aim is to capture the spatial relationships between vehicles and their surrounding environment. This is achieved by merging features from multiple modalities into one cohesive cross-modal representation. Diverging from a conventional attention-based framework, this fusion includes three main components: spatial embedding, cross-modal attention, and a regression head. The spatial embedding module first encodes the input features and maps them into a high-dimensional spatial space. Next, the cross-modal attention mechanism evaluates the spatial-contextual relationships among the various modalities and iteratively refines their representations. Ultimately, the regression head combines these refined representations to produce the high-demotion cross-modal features.

\textbf{Spatial Embedding.}
This module utilizes parameter-independent Gated Linear Units (GLUs) \cite{dauphin2017language} alongside the Gaussian Error Linear Unit (GELU) activation function \cite{hendrycks2016gaussian} to map the input vehicle features $\mathcal{F}_{v}$, HD map features $\mathcal{F}_{h}$, and dynamic features $\mathcal{F}_{d}$ into a high-dimensional embedding space on a frame-by-frame basis. This process ensures that each feature type undergoes specialized transformations, thereby maintaining the unique characteristics of each modality while capturing essential spatial relationships needed for cross-modal fusion. Correspondingly, it outputs the encoded high-demotion features $\mathcal{\hat{F}}_{v}, \mathcal{\hat{F}}_{h}$, and $\mathcal{\hat{F}}_{d}$, which serve as the foundational input for the subsequent cross-modal attention mechanism. The process can be formally expressed as follows:
\begin{equation}
\mathcal{\hat{F}}_{v} = \{\mathcal{\hat{F}}_{t},\mathcal{\hat{F}}_{s}\}=\phi_{\textit {GELU}}\left(\phi_{\textit {GLUs}} (\mathcal{F}_{v})\right), \,
\mathcal{\hat{F}}_{h} = \phi_{\textit{GELU}}\left(\phi_{\textit{GLUs}} (\mathcal{F}_{h})\right), \,
\mathcal{\hat{F}}_{d} = \phi_{\textit{GELU}}\left(\phi_{\textit{GLUs}} (\mathcal{F}_{d})\right)
\end{equation}

Here, \(\phi_{\textit{GLUs}}\) represents the GLUs. It presents a mechanism that regulates the information flow within the network, thereby improving the model's adaptability, which is mathematically represented as follows:
\begin{equation}
    \phi_{\textit{GLUs}}(\alpha) = ( \alpha W_a + b ) \odot \phi_{\textit{sigmoid}}( \alpha W_b + \hat{b})
\end{equation}
where \( \alpha \) denotes the input features, \( W_a \) and \( W_b \) are the learnable weight parameters associated with the GLUs layer, \( b\) and \( \hat{b} \) are the corresponding biases, while \( \odot \) is the element-wise multiplication.

\textbf{Cross-modal Attention.} 
As depicted in Figure \ref{cross} (b), the proposed cross-modal attention framework comprises \(N\) cross-modal attention blocks. Each block refines the spatial dependencies using cross-modal attention heads, followed by MLPs to perform non-linear mappings, thereby generating high-dimensional cross-modal features. To capture the spatial dependencies among various features effectively and ensure robust cross-modal fusion, we utilize two distinct channels: the target-spatial channel and the surrounding-spatial channel. These channels operate in parallel to compute the spatial contextual dependencies between the target vehicle and the environment, as well as between the surrounding vehicles and the environment. Specifically, in the $i$-th cross-modal attention block, the encoded features $\mathcal{\hat{F}}_{s}, \mathcal{\hat{F}}_{t}$, and $\mathcal{\hat{F}}_{h}$ are projected into two sets of query, key, and value vectors, termed the target-spatial group $\{Q_t^i, K_t^i, V_t^i\}$ and the surrounding-spatial group $\{Q_s^i, K_s^i, V_s^i\}$, respectively. As illustrated in Figure \ref{cross} (b), additional dynamic features $\mathcal{\hat{F}}_{d}$ are appended to the end of the encoded vehicle features $\mathcal{\hat{F}}_{v}$ and HD map features $\mathcal{\hat{F}}_{h}$ to improve the model’s capability in capturing the dynamics behavior of vehicles. This can be mathematically expressed as follows:
\begin{equation}
\left\{\begin{array}{l}
{Q}_t^i={W}^{Q}_t (\phi_{\textit{MLPs}}(\mathcal{\hat{F}}_{v})+\mathcal{\hat{F}}_{d})\\

{K}_t^i={W}^{K}_t (\phi_{\textit{MLPs}}(\mathcal{\hat{F}}_{h})+\mathcal{\hat{F}}_{d})\\

{V}_t^i={W}^{V}_t (\phi_{\textit{MLPs}}(\mathcal{\hat{F}}_{t}))
\end{array}\right.
\end{equation}
\begin{equation}
\left\{\begin{array}{l}
{Q}_s^i={W}^{Q}_s (\phi_{\textit{MLPs}}(\mathcal{\hat{F}}_{v})+\mathcal{\hat{F}}_{d})\\

{K}_s^i={W}^{K}_s (\phi_{\textit{MLPs}}(\mathcal{\hat{F}}_{h})+\mathcal{\hat{F}}_{d})\\

{V}_s^i={W}^{V}_s (\phi_{\textit{MLPs}}(\mathcal{\hat{F}}_{s}))
\end{array}\right.
\end{equation}
where ${W}^{Q}_t, {W}^{K}_t, {W}^{V}_t, {W}^{Q}_s, {W}^{K}_s, {W}^{V}_s$ are all the learnable weights.

Moreover, the output of the \(i\)-th cross-modal attention head, $\mathcal{H}^i_{c}$, is obtained by summing the target-spatial vectors $\mathcal{H}_{t}^i$ and surrounding-spatial $\mathcal{H}_{s}^i$ vectors, where $i \in [1,N]$. Formally,
\begin{flalign}
{\mathcal{H}^i_{c}} &=\mathcal{H}_{t}^i + \mathcal{H}_{s}^i \\\nonumber&=\underbrace{{ \phi_{\textit {softmax}}} \left( \phi_{\textit {MLPs}} \bigg(\frac{{Q}^i_t ({K}^i_t)^{\top}}{\sqrt{d_t}} \bigg)\right) \phi_{\textit {MLPs}} ({V}^i_t)}_{\mathcal{H}_{t}^i} + 
\underbrace{ \phi_{\textit {softmax}} \left(\phi_{\textit {MLPs}}\bigg(\frac{{Q}^i_s ({K}^i_s)^{\top}}{\sqrt{d_s}}\bigg)\right) \phi_{\textit {MLPs}}({V}^i_s)}_{\mathcal{H}_{s}^i}
\end{flalign}
where $\phi_{\textit{softmax}}$ is the softmax activation function. $d_t$ and $d_s$ represent the dimensionalities of the key vectors, which are applied to scale the results and avoid excessively large values in the matrix. 

Subsequently, all the outputs of the cross-modal attention block are processed with the normalization layer, followed by MLPs. The accumulated results yield the high-dimension hidden embedding $\mathcal{{H}}_{c}$. This can be represented as follows:
\begin{flalign}
\mathcal{{H}}_{e} = \sum_{i=1}^{N} \big( \phi_{\textit {MLPs}}(\phi_{\textit {LN}}( {\mathcal{H}^i_{c}}+\mathcal{\hat{F}}_{h})) + ({\mathcal{H}^i_{c}}+\mathcal{\hat{F}}_{h}) \big )
\end{flalign}

\textbf{Regression Head.}
Following the cross-modal attention, we integrate a regression head at the top of this component. The high-dimensional hidden embedding $\mathcal{{H}}_{e}$ is fed into this regression head, which consists of a 4-layer MLP network with ReLU activation functions, except for the final layer, resulting in the high-dimensional cross-modal features $\mathcal{F}_{c}$ that encapsulate rich spatial context.

\subsubsection{Spatial-temporal Encoder}
This encoder aims to generate interaction features by capturing both the temporal dependencies from vehicle features $\mathcal{F}_{v}$ and the spatial dependencies from cross-modal features $\mathcal{F}_{c}$. It starts by applying a 2-layer Gated Recurrent Unit network to efficiently extract frame-wise temporal feature maps. In parallel, a two-layer Graph Convolutional Network is employed to model the spatial relationships embedded within the cross-modal features $\mathcal{F}_{c}$, allowing the network to fully comprehend the spatial dependencies between vehicles and their environments. The outputs of these networks are then fused and passed through a transformer encoder, which updates and refines the intricate spatial-temporal interactions. This enables the model to produce high-level interaction features $\mathcal{F}_{i}$ that encapsulate both temporal evolution and spatial dependencies.

\subsection{Multi-modal Decoder}
The multi-modal decoder is designed to address the inherent uncertainty in trajectory prediction by evaluating multiple possible maneuvers and their respective probabilities. To achieve this, we integrate the interaction features \(\mathcal{F}_{i}\) and dynamic features \(\mathcal{F}_{d}\), reshaping them with MLPs to form a composite token. This token is then processed by a GLU, which systematically analyzes the dynamics real-time variations captured in the dynamic features while capturing the spatial-temporal dependencies in the interaction features. Finally, the output of the GLU is passed through MLPs to generate the predicted trajectories \( \bm{Y}^{t_c+1:t_c+t_f}_0 \) and the corresponding maneuver probability distributions. Overall, this multi-modal structure not only allows for different trajectory predictions but also quantifies the confidence for each prediction, providing a robust basis for dealing with uncertainty in decision-making.

\section{Experiments}\label{Experiment}

\subsection{Experiment Setup} 
\textbf{Datasets.} This study utilizes four well-established real-world datasets, carefully selected to rigorously evaluate the robustness and accuracy of our proposed DEMO model across diverse traffic conditions. These datasets include NGSIM \cite{deo2018convolutional} and HighD \cite{8569552}, both providing high-resolution highway trajectory data; MoCAD \cite{liao2024bat}, focusing on urban and campus environments with unique right-hand driving dynamics; nuScenes \cite{caesar2020nuscenes}, renowned for its complex multi-vehicle traffic scenes involving pedestrians, riders, and vehicles in unstructured environments. These datasets collectively cover a wide array of driving environments, from congested urban streets to high-speed highways, including intersections, campuses, and multi-vehicle road scenarios, ensuring a comprehensive validation of DEMO's ability to predict vehicle trajectories under various challenging driving conditions.

\textbf{Dataset Segmentations.} For the NGSIM, MoCAD, and HighD datasets, we adopt a 3-second past horizon $t_p$ and a 5-second prediction horizon $t_f$. For the nuScenes dataset, we use a 2-second past horizon $t_p$ and a 6-second prediction horizon $t_f$. Following the approaches in \cite{chen2022intention, chandra2020forecasting}, the short-term prediction horizon $t_s$ is defined as 2 seconds.

\textbf{Implementation Details.} DEMO is developed, trained, and evaluated using the PyTorch framework. We employ the AdamW optimizer \cite{loshchilov2017decoupled} to manage weight decay, with an initial learning rate of 0.001. During training, the learning rate is adjusted using the CosineAnnealingLR schedule \cite{loshchilov2022sgdr}. In the following experiments, all evaluations are conducted on an Nvidia A40 48GB GPU, except for the inference time comparison, which is performed on an Nvidia RTX 3090 GPU. To maintain consistency with the models \cite{liao2024bat,chen2022intention} developed based on the NGSIM, MoCAD, and HighD datasets, HD map data is excluded as an input during evaluations on these datasets. Consequently, the HD map features $\mathcal{F}_{h}$ in the cross-modal fusion are replaced with vehicle features $\mathcal{F}_{v}$.

\textbf{Evaluation Metrics.}
In this study, three widely recognized metrics are employed: Root Mean Square Error (RMSE) \cite{liao2024mftraj}, Minimum Average Displacement Error (\textit{minADE}), and Minimum Final Displacement Error (\textit{minFDE}) \cite{moon2025visiontrap}. To align with benchmarks \cite{liao2024less, chen2022intention} in the field, including the NGSIM, MoCAD, and HighD datasets, we use RMSE calculated every 1 second as the evaluation metric, facilitating comparison of different models' predictive accuracy over short-term and long-term prediction horizons, respectively. Additionally, $\textit{minADE}$ and $\textit{minFDE}$ are employed to assess the performance on the nuScenes dataset.

\textbf{Loss Functions.}
To ensure optimal performance, we train the DEMO model using a weighted combination of several loss functions. For the DynCVAE, the loss function includes the KL divergence loss, denoted as $\mathcal{L}_\textit{KL}$, and a dynamics-informed loss, $\mathcal{L}_\textit{DI}$. To further guide the model in correctly identifying the driver's intended maneuvers, we incorporate a cross-entropy loss function, $\mathcal{L}_\textit{CE}$. This provides supervision for maneuver classification, ensuring the model interprets the driver's behavior accurately. In addition to maneuver classification, the accuracy of the predicted trajectory is enforced through an accuracy loss function, $\mathcal{L}_\textit{AC}$. This loss function is adapted based on the dataset being used, aligning model performance with specific benchmark requirements. For example, for datasets such as NGSIM, MoCAD, and HighD, the accuracy loss is MSE and negative log-likelihood loss \cite{chen2022intention}. In contrast, in the nuScenes dataset, we use a weighted sum of $\textit{minADE}$ and $\textit{minFDE}$ as the accuracy loss.

\begin{table}[htbp]
  \centering
  \caption{Evaluation results of the proposed DEMO model and other baselines over the short-term prediction horizon. \textbf{Bold} and \underline{underlined} values represent the best and second-best performance in each category. Cases marked as ('-') indicate unspecified values.}  \setlength{\tabcolsep}{9mm}
  \resizebox{0.96\linewidth}{!}{
    \begin{tabular}{c|cc|cc|cc}
    \toprule
    \multicolumn{1}{c}{\multirow{2}[3]{*}{Model}} & \multicolumn{2}{c}{NGSIM} & \multicolumn{2}{c}{MoCAD} & \multicolumn{2}{c}{HighD}\\
\cmidrule{2-7}   \multicolumn{1}{c}{}   & 1     & 2         & 1     & 2      & 1     & 2          \\
    \hline
           S-GAN \cite{gupta2018social} & 0.57  & 1.32    & 1.69  & 2.25    & 0.30  & 0.78  \\
           CS-LSTM \cite{deo2018convolutional} & 0.61  & 1.27    & 1.45  & 1.98    & 0.22  & 0.61  \\
           CS-LSTM(M) \cite{deo2018convolutional}& -  & -   & 1.49  & 2.07   & 0.23  & 0.65  \\
           NLS-LSTM \cite{messaoud2019non} & 0.56  & 1.22    & 0.96  & 1.27   & 0.20  & 0.57  \\
           CF-LSTM \cite{xie2021congestion} & 0.55  & 1.10    & 0.72  & 0.91   & 0.18  & 0.42   \\
            iNATran \cite{chen2022vehicle} & \underline{0.39}  &0.96    & -  & -  & \textbf{0.04}  & \textbf{0.05}  \\
            BAT \cite{liao2024bat} & - & -  & {0.35}  & {0.74}  & - & -  \\
           MHA-LSTM \cite{messaoud2021attention} & 0.41  & 1.01  & 1.25  & 1.48  & 0.19  & 0.55 \\
           STDAN \cite{chen2022intention} & \underline{0.39}  & 0.96    & 0.62  & 0.85   & 0.19  & 0.27 \\
           DACR-AMTP \cite{cong2023dacr}& 0.57  & 1.07    & -  & -  & 0.10  & {0.17}  \\ 

            WSiP \cite{wang2023wsip} & 0.56  & 1.23   & 0.70  & 0.87   & 0.20  & 0.60 \\
           GaVa \cite{liao2024human} & {0.40}  & {0.94}    & -  & -  & 0.17  & 0.24    \\
    \hline
   \textbf{DEMO} & \textbf{0.36} & \textbf{0.86} & \textbf{0.26} & \textbf{0.41} & \underline{0.06} & \underline{0.14} \\
   \textbf{DEMO (50\%)} & \underline{0.39}  & \underline{0.91}  &\underline{0.33}  & \underline{0.64}   & {0.07}  & 0.17  \\
    \bottomrule
    \end{tabular}%
    }
  \label{short_term}%
\end{table}%

\begin{table}[htbp]
  \centering
  \caption{Evaluation results of the proposed DEMO model and other baselines over the long-term prediction horizon. \textbf{Bold} and \underline{underlined} values represent the best and second-best performance in each category. Cases marked as ('-') indicate unspecified values.}  \setlength{\tabcolsep}{6mm}
  \resizebox{0.96\linewidth}{!}{
    \begin{tabular}{c|ccc|ccc|ccc}
    \toprule
    \multicolumn{1}{c}{\multirow{2}[3]{*}{Model}} & \multicolumn{3}{c}{NGSIM} & \multicolumn{3}{c}{MoCAD} & \multicolumn{3}{c}{HighD}\\
\cmidrule{2-10}   \multicolumn{1}{c}{}   & 3     & 4     & 5     & 3     & 4     & 5    & 3     & 4     & 5     \\
    \hline
           S-GAN \cite{gupta2018social}  & 2.22  & 3.26  & 4.20 & 3.30   & 3.89  & 4.69  & 1.46 & 2.34  & 3.41\\
           CS-LSTM \cite{deo2018convolutional} & 2.09 & 3.10  & 4.37   & 2.94& 3.56  & 4.49   & 1.24& 2.10  & 3.27\\
           CS-LSTM(M) \cite{deo2018convolutional}& -  & -  & -   & 3.02 & 3.62  & 4.53   & 1.29  & 2.18  & 3.37 \\
           NLS-LSTM \cite{messaoud2019non}  & 2.02 & 3.03  & 4.30    & 2.08  & 2.86  & 3.93 & 1.14  & 1.90  & 2.91 \\
           CF-LSTM \cite{xie2021congestion}  & 1.78  & 2.73  & 3.82   & 1.73& 2.59  & 3.44   & 1.07 & 1.72  & 2.44 \\
            iNATran \cite{chen2022vehicle}   &1.61 & 2.42  & 3.43 & -  & -  & -  & \textbf{0.21} & \underline{0.54}  & 1.10\\
            BAT \cite{liao2024bat} & - & -  & -  & {1.39} & \underline{2.19} &\underline{2.88}  & - & -  & -\\
           MHA-LSTM \cite{messaoud2021attention} & 1.74 & 2.67  & 3.83    & 2.57 & 3.22  & 4.20   & 1.10 & 1.84  & 2.78 \\
           STDAN \cite{chen2022intention}  & 1.61& 2.56 & 3.67   & 1.62  & 2.51  & 3.32  & 0.48  & 0.91  & 1.66 \\
           DACR-AMTP \cite{cong2023dacr}  & 1.68 & 2.53  & 3.40  & -  & -  & -  & 0.31  & \underline{0.54}  & \underline{1.01}\\ 

            WSiP \cite{wang2023wsip}  & 2.05 & 3.08  & 4.34     & 1.70 & 2.56  & 3.47  & 1.21 & 2.07  & 3.14\\
           GaVa \cite{liao2024human}   & \underline{1.52} & \underline{2.24}  & \underline{3.13}  & -  & -  & -  & 0.42 & 0.86  & 1.31 \\
    \hline
   \textbf{DEMO}  & \textbf{1.48} & \textbf{ 2.10 } & \textbf{ 2.88 }   & \textbf{1.03} & \textbf{ 1.78 } & \textbf{ 2.67 }   & \underline{0.25} & \textbf{0.44} & \textbf{0.70} \\
   \textbf{DEMO (50\%)}  & 1.60  & 2.35 & 3.36   & \underline{1.28} & 2.27 & 3.18  & 0.33 & 0.70 & 1.08  \\
    \bottomrule
    \end{tabular}%
    }
  \label{long_term}%
\end{table}%

\begin{table*}[!ht]
\centering
\caption{Performance comparison of various models on nuScenes dataset at long-term prediction horizon (6-second). Metrics include $\textit{minADE}_{10}$, $\textit{minADE}_5$, $\textit{minADE}_1$ and $\textit{minFDE}_1$.\textbf{Bold} and \underline{underlined} values represent the best and second-best performance in each category. Cases marked as ('-') indicate unspecified values. }
\setlength{\tabcolsep}{6mm}
\resizebox{0.7\textwidth}{!}{
\begin{tabular}{ccccccc}
 \bottomrule
\multirow{2}{*}{Model} &  \multirow{2}{*}{$\textit{minADE}_{10}\downarrow$ } & \multirow{2}{*}{$\textit{minADE}_5\downarrow$}  & \multirow{2}{*}{$\textit{minADE}_1\downarrow$} 
& \multirow{2}{*}{$\textit{minFDE}_1\downarrow$} \\
\\
\hline
Const vel and yaw &  4.61 & 4.61 & 4.61 & 11.21 \\
Physics Oracle& 3.70 & 3.70 & 3.70 & 9.09 \\
MultiPath \cite{chai2020multipath} &  1.14 & 1.44 & \underline{3.16} & 7.69 \\
CoverNet \cite{phan2020covernet}  &    1.92 & 2.62 & - & 11.36 \\
DLow-AF \cite{yuan2020dlow} &   1.78 & 2.11 & - & - \\
Trajectron++ \cite{salzmann2020trajectron++} &   1.51 & 1.88 & - & 9.52 \\
LDS-AF \cite{ma2021likelihood} &    1.65 & 2.06 & - & - \\
MHA-JAM \cite{messaoud2021trajectory} &  1.24 & 1.81 & 3.69 & 8.57 \\
LaPred \cite{kim2021lapred} & 1.12 & 1.53 & 3.51 & 8.12 \\
STGM \cite{zhong2022stgm} &  2.34 & - & 3.21 & 9.62\\ 
GoHome \cite{gilles2022gohome} &  1.15 & 1.42 & - & \underline{6.99} \\
PGP \cite{deo2022multimodal} &  \underline{1.00} & 1.30 & - & - \\
ContextVAE \cite{xu2023context} &   - & 1.59 & 3.54 & 8.24 \\
Q-EANet\cite{chen2024q} &   1.02 & \textbf{1.18} & - & -\\
EMSIN \cite{ren2024emsin} &    - & 1.77 & 3.56 & - \\
AFormer-FLN\cite{xu2024adapting} &1.32 & 1.83 & - & - \\
E-\(V^2\)-Net-SC\cite{wong2024socialcircle} &  1.13 & 1.44 & - & - \\
SeFlow \cite{zhang2025seflow} &  \textbf{0.98} & 1.38 & - & 7.89 \\
\hline
\textbf{DEMO} &  {1.04} & \underline{1.20} & \textbf{2.99} & \textbf{6.90} \\
\textbf{DEMO (50\%)} &  {1.14} & {1.32} & {3.24} & {7.28} \\
\toprule
\end{tabular}}
\vspace{-6pt}
\label{nuscenes_result}
\end{table*}

\subsection{Experimental Results} 
To thoroughly assess DEMO’s performance, we approach the evaluation from three key perspectives: short-term horizon prediction accuracy, long-term horizon prediction accuracy, and inference speed. Each of these metrics offers unique insights into the model's capability to handle various aspects of trajectory prediction, essential for real-world applications in autonomous driving. Furthermore, to rigorously evaluate DEMO's adaptability and efficiency, we conducted training using only half of the available data and the results is denoted as DEMO (50\%).

\textbf{Comparsion Analysis on Short-term Horizon.} Table \ref{short_term} compares the performance of the DEMO model with other models over the short-term prediction horizon. On the NGSIM dataset, known for its dense highway scenarios, DEMO demonstrates significant improvements. Notably, in terms of RMSE over a 2-second horizon, we observe an improvement of 8.5\%. Even with only 50\% of the training dataset available, DEMO (50\%) achieves SOTA performance. These results indicate that DEMO effectively captures vehicle motion dynamics, significantly enhancing short-term prediction accuracy in dense highway scenarios. In our evaluation on the MoCAD dataset, characterized by complex urban streets and a right-hand driving system, DEMO demonstrates significant advancements over SOTA baselines, with impressive improvements of 25.7\% and 44.6\% for the 1- and 2-second short-term prediction horizon. Moreover, even with only 50\% of the training data, DEMO (50\%) surpasses all previous SOTA baselines in short-term prediction horizons, highlighting the contribution of combining dynamic models with deep learning algorithms for enhanced short-term prediction accuracy. Although iNATran \cite{chen2022vehicle} achieved top results at the 1- and 2-second horizons on the HighD dataset, our DEMO model secures second place. Additionally, DEMO (50\%) outperforms all baselines except iNATran, further underscoring the superiority of the DEMO model in short-term prediction.

\textbf{Comparsion Analysis on Long-term Horizon.} Table \ref{long_term} presents the evaluation results conducted over a long-term horizon on the NGSIM, MoCAD, and HighD datasets. The DEMO model's superior performance across all datasets, including highway, urban, and campus road scenarios, can be attributed to the Interaction Learning Stage's meticulous modeling of interactions within the environment. In particular, at the 4-second prediction horizon, improvements of 6.3\%, 18.7\%, and 18.5\% are observed. For the 5-second horizon, improvements reach 8.0\%, 7.3\%, and 30.7\%, respectively. Table \ref{nuscenes_result} presents the long-term prediction performance of different models on the nuScenes dataset. The evaluation results underscore DEMO’s impressive ability to navigate complex urban streets and intricate intersection scenarios. Notably, in terms of $\textit{minADE}_1$ and $\textit{minFDE}_1$, we observe improvements of at least 5.4\% and 1.3\%, respectively. These results highlight the capability of the Interaction Learning Stage within the DEMO model to effectively model interaction patterns in complex scenarios. While DEMO (50\%) performs less impressively on long-term prediction horizons compared to its short-term horizon results. We attribute this to the fact that deep learning models rely heavily on large volumes of historical data to model interaction patterns within the scene; thus, as data availability decreases, performance naturally declines.

\begin{table}[tbp]
\centering
\setlength{\tabcolsep}{9mm}
\caption{Comparison of inference time on nuScenes dataset.}
\resizebox{0.5\linewidth}{!}
{
\begin{tabular}{cc}
\bottomrule
\text{Model} & Inference Time (ms)  \\
\midrule
P2T \cite{deo2020trajectory} & 116  \\
Trajectron++ \cite{salzmann2020trajectron++} &     38   \\
MultiPath \cite{chai2020multipath} &   87   \\
AgentFormer \cite{yuan2021agentformer} &   107   \\
PGP \cite{deo2022multimodal} & 215  \\
LAformer \cite{liu2024laformer} & 115 \\
VisionTrap \cite{moon2025visiontrap} & 53\\
 \midrule
\textbf{DEMO} & 15  \\
\toprule
\end{tabular}}
\label{table:performance_inference}
\end{table}
\textbf{Inference Speed Evaluation.}
Inference speed is a crucial metric for evaluating trajectory prediction models. To measure inference speed, we compute the average time taken by various models to predict trajectories for 12 samples. Table \ref{table:performance_inference} presents a comparison of the inference times across different models. It is important to note that while the other models are evaluated on an RTX 3090 Ti GPU \cite{moon2025visiontrap}, DEMO is tested on the RTX 3090 GPU. We observed that our proposed model significantly outperforms the others in terms of inference speed, highlighting its practicality for real-world applications.

In summary, DEMO demonstrates exceptional accuracy, resource efficiency, and adaptability across challenging driving environments, establishing it as a highly reliable solution for trajectory prediction. By consistently outperforming SOTA models across multiple datasets—including highways, urban streets, and intersections—DEMO proves its robustness and predictive precision. Notably, its fast inference speed underscores its suitability for real-time applications. These attributes make DEMO a versatile and scalable solution, enhancing the safety of AVs.

\begin{table}[tbp]
  \centering
  \caption{Ablation setting in this study. \ding{52} denotes the corresponding component is maintained during evaluation, while \ding{56} denotes the corresponding component is excluded during evaluation.}
  \setlength{\tabcolsep}{7mm}
  \resizebox{0.8\linewidth}{!}{
            \begin{tabular}{cccccccc}
                \toprule
                \multirow{2}[4]{*}{Components} & \multicolumn{6}{c}{Ablation Methods} \\
                \cmidrule{2-7}          & A     & B     & C     & D     & E  &F  \\
                \midrule
                Dynamic Bicycle Model & \ding{56} & \ding{56} & \ding{52} & \ding{52} & \ding{52} & \ding{52} \\
                DynCVAE  & \ding{52} & \ding{56} & \ding{52}& \ding{52}  & \ding{52} & \ding{52}  \\
                 Temporal Encoder& \ding{52} & \ding{52} &  \ding{56} &\ding{52} & \ding{52}  & \ding{52} \\
                 Cross-modal Fusion & \ding{52} & \ding{52} & \ding{52} & \ding{56} & \ding{52}  & \ding{52}\\
                  Spatial-temporal Encoder & \ding{52} & \ding{52} & \ding{52} & \ding{52} & \ding{56}  & \ding{52}\\
                 Multi-modal Decoder & \ding{52} & \ding{52} & \ding{52} & \ding{52}& \ding{52}  & \ding{56}\\
                \bottomrule
            \end{tabular}}%
  \label{ab2_strategy_nuScenes}%
\end{table}%

\begin{table}[t]
  \centering
  \caption{Ablation results on NGSIM, MoCAD, and HighD datasets over the short-term horizon. Metric: RMSE}
   \setlength{\tabcolsep}{5mm}
  \resizebox{\linewidth}{!}{
        \begin{tabular}{c|cccccccc}
        \toprule
        \multicolumn{1}{c}{\multirow{2}[3]{*}{Dataset}} & \multirow{2}[3]{*}{Time (s)} & \multicolumn{6}{c}{Ablation Methods} \\
    \cmidrule{3-9}     \multicolumn{1}{c}{}     &  & A     & B     & C  &D   & E   & F &\textbf{DEMO}\\
        \midrule
        \multirow{2}[1]{*}{NGSIM} 
        & 1 & 0.42$_{\downarrow{17\%}}$ & 0.58$_{\downarrow{61\%}}$ & 0.46$_{\downarrow{28\%}}$  & 0.54$_{\downarrow{50\%}}$&0.43$_{\downarrow{19\%}}$ & 0.45$_{\downarrow{25\%}}$ & 0.36\\%
        & 2 & 0.90$_{\downarrow{5\%}}$ & 1.22$_{\downarrow{42\%}}$ & 1.03$_{\downarrow{20\%}}$ & 1.10$_{\downarrow{28\%}}$ &1.04$_{\downarrow{21\%}}$ & 1.07$_{\downarrow{24\%}}$ &0.86\\%
        \midrule
        \multirow{2}[1]{*}{MoCAD} 
        & 1 & 0.30$_{\downarrow{15\%}}$ & 0.43$_{\downarrow{65\%}}$ & 0.36$_{\downarrow{38\%}}$ & 0.40$_{\downarrow{54\%}}$&0.31$_{\downarrow{30\%}}$ & 0.33$_{\downarrow{27\%}}$&0.26 \\
        & 2 & 0.49$_{\downarrow{20\%}}$ & 0.70$_{\downarrow{71\%}}$ & 0.59$_{\downarrow{44\%}}$ & 0.64$_{\downarrow{56\%}}$&0.55$_{\downarrow{34\%}}$ & 0.60$_{\downarrow{46\%}}$&0.41\\
        \midrule
        \multirow{2}[1]{*}{HighD} 
        & 1 & 0.07$_{\downarrow{33\%}}$ & 0.13$_{\downarrow{117\%}}$ & 0.09$_{\downarrow{50\%}}$ & 0.11$_{\downarrow{83\%}}$ &0.07$_{\downarrow{17\%}}$   & 0.08$_{\downarrow{33\%}}$ &0.06\\
        & 2 & 0.18$_{\downarrow{29\%}}$ & 0.31$_{\downarrow{121\%}}$ & 0.25$_{\downarrow{79\%}}$ & 0.29$_{\downarrow{107\%}}$ &0.21$_{\downarrow{50\%}}$  & 0.25$_{\downarrow{79\%}}$ &0.14\\%
        \bottomrule
        \end{tabular}}%
  \label{ab2_short}%
\end{table}%

\begin{table}[t]
  \centering
  \caption{Ablation results on NGSIM, MoCAD, and HighD datasets over the long-term horizon. Metric: RMSE}
   \setlength{\tabcolsep}{7mm}
  \resizebox{\linewidth}{!}{
        \begin{tabular}{c|cccccccc}
        \toprule
        \multicolumn{1}{c}{\multirow{2}[3]{*}{Dataset}} & \multirow{2}[3]{*}{Time (s)} & \multicolumn{6}{c}{Ablation Methods} \\
    \cmidrule{3-9}     \multicolumn{1}{c}{}     &  & A     & B     & C  &D   & E   & F &\textbf{DEMO}\\
        \midrule
        \multirow{3}[1]{*}{NGSIM} 
        & 3 & 1.52$_{\downarrow{3\%}}$ & 1.60$_{\downarrow{8\%}}$ & 1.54$_{\downarrow{4\%}}$ & 1.79$_{\downarrow{21\%}}$ &1.60$_{\downarrow{8\%}}$  & 1.55$_{\downarrow{5\%}}$ &1.48\\%
        & 4 & 2.17$_{\downarrow{3\%}}$ & 2.23$_{\downarrow{6\%}}$ & 2.32$_{\downarrow{10\%}}$ & 2.56$_{\downarrow{22\%}}$&2.36$_{\downarrow{12\%}}$ & 2.41$_{\downarrow{15\%}}$ &2.10\\
        & 5 & 2.94$_{\downarrow{2\%}}$ & 3.10$_{\downarrow{8\%}}$ & 3.16$_{\downarrow{10\%}}$ & 3.43$_{\downarrow{19\%}}$&3.18 $_{\downarrow{10\%}}$& 3.24$_{\downarrow{13\%}}$ &2.88\\
        \midrule
        \multirow{3}[1]{*}{MoCAD} 
        & 3 & 1.07$_{\downarrow{4\%}}$ & 1.12$_{\downarrow{9\%}}$ & 1.24$_{\downarrow{20\%}}$ & 1.36$_{\downarrow{32\%}}$ &1.20$_{\downarrow{17\%}}$  & 1.32$_{\downarrow{28\%}}$ &1.03\\%
        & 4 & 1.81$_{\downarrow{2\%}}$ & 1.90$_{\downarrow{7\%}}$ & 2.18$_{\downarrow{22\%}}$ & 2.24$_{\downarrow{26\%}}$ &2.11$_{\downarrow{19\%}}$ & 2.15$_{\downarrow{21\%}}$&1.78 \\
        & 5 & 2.71$_{\downarrow{1\%}}$ & 2.79$_{\downarrow{4\%}}$ & 2.88$_{\downarrow{8\%}}$ & 3.12$_{\downarrow{17\%}}$ &2.86$_{\downarrow{7\%}}$  & 2.94$_{\downarrow{10\%}}$ &2.67\\
        \midrule
        \multirow{3}[1]{*}{HighD} 
        & 3 & 0.29$_{\downarrow{16\%}}$ & 0.42$_{\downarrow{68\%}}$ & 0.48$_{\downarrow{92\%}}$ & 0.54$_{\downarrow{116\%}}$ &0.35$_{\downarrow{40\%}}$  & 0.47$_{\downarrow{88\%}}$ &0.25\\%
        & 4 & 0.53$_{\downarrow{20\%}}$ & 0.58$_{\downarrow{32\%}}$ & 0.65$_{\downarrow{48\%}}$ & 0.78$_{\downarrow{77\%}}$ &0.60$_{\downarrow{36\%}}$  & 0.67$_{\downarrow{52\%}}$ &0.44\\
        & 5 & 0.77$_{\downarrow{10\%}}$ & 0.82$_{\downarrow{17\%}}$ & 0.90$_{\downarrow{29\%}}$ & 1.05$_{\downarrow{50\%}}$ &0.89$_{\downarrow{27\%}}$   & 0.95$_{\downarrow{36\%}}$ &0.70\\
        \bottomrule
        \end{tabular}}%
  \label{ab2_long}%
\end{table}%

\subsection{Ablation Studies}
To explore the contribution of each component to DEMO's prediction performance and to better understand the mechanism of our DEMO in short-term and long-term prediction, we conduct this ablation study. As shown in Table \ref{ab2_strategy_nuScenes}, the specific configurations for each ablation method are as follows: Method A replaces the Dynamic Bicycle Model with a simpler kinematics model, reducing the model's sensitivity to instantaneous dynamics. Method B removes both the DynCVAE and the Dynamic Bicycle Model, discarding the entire Dynamics Learning Stage. Method C discards the temporal encoder, impairing the model’s ability to perceive temporal dependencies. Method D removes the cross-modal fusion framework, which is essential for the model's comprehensive understanding of spatial dependencies. Method E discards the spatial-temporal encoder, which enables the model to capture the interaction between spatial and temporal dimensions. Finally, Method F omits the proposed multi-modal decoder module.

Table \ref{ab2_short} presents the results of ablation experiments within the short-term horizon. It is evident that Method B result in the most substantial accuracy decline, with reductions of at least 42\%. This outcome underscores the impact of vehicle motion dynamics on short-term prediction. The ablation experiments for the long-term horizon were conducted on the NGSIM, MoCAD, HighD, and nuScenes datasets. Table \ref{ab2_long} displays results from the first three datasets, where we observe that modules within the Dynamics Learning Stage have a less pronounced impact on long-term prediction accuracy compared to short-term prediction. In contrast, removing any module from the Interaction Learning Stage leads to significant accuracy degradation. The exclusion of the cross-modal fusion framework, in particular, has the largest impact, with accuracy drops of at least 17\%. Table \ref{ab2_nuscenes} shows the experimental results on the nuScenes dataset, where a similar trend can be observed. Each ablation variant exhibits a varying degree of performance decline, emphasizing the importance of each module within the proposed DEMO model for achieving high-accuracy predictions.

\begin{table}[tbp]
  \centering
  \caption{Ablation results on nuScenes at the long-term horizon (6-second horizon).}
  \setlength{\tabcolsep}{6mm}
  \resizebox{0.8\linewidth}{!}
  {
            \begin{tabular}{ccccccccc}
            \toprule
            \multirow{2}[4]{*}{Metrics} & \multicolumn{7}{c}{Ablation Methods} \\
        \cmidrule{2-8}          & A     & B     & C     & D     & E  &F & \textbf{DEMO}\\
            \midrule
          $\text{mADE}_{10}$ & 1.07 & 1.25 & 1.13 & 1.31 & 1.12 & 1.05 & 1.04\\
             $\text{mADE}_5$ & 1.28 & 1.46 & 1.38 & 1.47 & 1.35 & 1.11 & 1.20\\
             $\text{mADE}_1$ & 3.05 & 3.23 & 3.14 & 3.31 & 3.21 & 3.19 & 2.99\\
             $\text{mFDE}_1$ & 6.94 & 7.22 & 7.12 & 7.40 & 7.18 & 7.17 & 6.90\\
            \bottomrule
            \end{tabular}}%
  \label{ab2_nuscenes}%
\end{table}%

\begin{figure}[tbp]
  \centering
\includegraphics[width=0.9\linewidth]{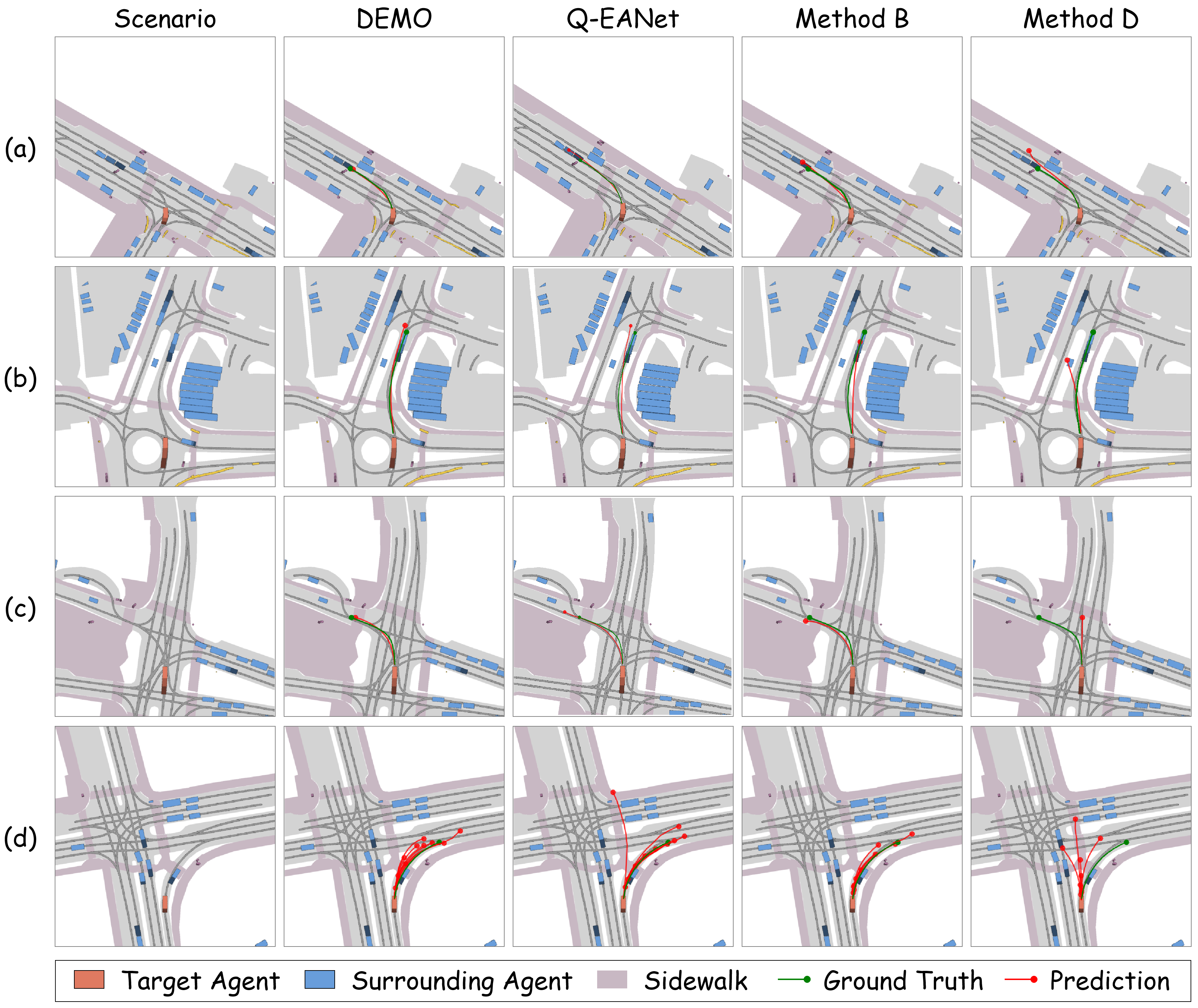} 
  \caption{Qualitative comparison of various models on nuScenes dataset. We select some representative scenarios including T-intersection (a), roundabout (b), interchange (c), and intersection (d). Panels (a), (b), and (c) visualize the most probable predictions of each model alongside the ground truth. Panel (d) visualizes the multi-modal prediction results in comparison with the ground truth. }
  \label{nu_keshihua} 
\end{figure}

\subsection{Qualitative Results}
Figure \ref{nu_keshihua} presents a visualization comparison of DEMO and two ablation variants, Method B and Method D, against the SOTA model Q-EANet \cite{chen2024q} across four different urban scenarios in the nuScenes dataset. Scenario (a) illustrates the prediction performance of different models for a target vehicle making a left turn at a T-intersection. We observe that DEMO demonstrates a clear advantage in the long-term prediction horizon compared to Q-EANet and Method D. This not only highlights the superior collective predictive performance of DEMO but also underscores the critical role of cross-modal fusion in enhancing long-term predictions. Moreover, scenario (b) showcases the prediction performance of different models for a target vehicle navigating a roundabout. Only DEMO’s predicted trajectory aligns closely with the ground truth, while the other models fail to achieve optimal results in both short-term and long-term predictions. Specifically, Method B, which removes the Dynamics Learning Stage, performs poorly in the short-term prediction horizon. In addition, Scenario (c) illustrates the prediction performance for a target vehicle on an interchange. In this complex road environment, Method D, which omits the cross-modal fusion, incorrectly interprets the driver’s intended maneuver, leading to significant prediction errors. Finally, scenario (d), highlights the multi-modal prediction capabilities of the models. We observe that both Q-EANet and Method D overlook the fact that the target vehicle is already in a right-turn-only lane, resulting in predictions that include straight or left-turn trajectories. This demonstrates the critical role of our cross-modal fusion framework in effectively perceiving accurate driver intention.

\section{Conclusion}\label{Conclusion}
In this study, we propose the DEMO model, consisting of two stages, namely Dynamics Learning Stage and Interaction Learning Stage, to confront the distinct challenges in short-term and long-term trajectory prediction tasks. In particular, we establish the Dynamics Learning Stage to extract precise vehicle motion dynamics to ensure accurate short-term prediction. This stage involves DynCVAE and Dynamic Bicycle Model, which work synergistically to capture dynamic features over the short term. As for the accurate interaction modeling, we introduce the Interaction Learning Stage. This stage involves a temporal encoder and cross-modal fusion framework to enable the DEMO to perceive temporal dependencies and spatial dependencies effectively, thus facilitating the generation of spatial-temporal interaction features. To the best of our knowledge, this is the pioneering approach to address short-term and long-term prediction with the fusion of vehicle dynamics models and deep learning algorithms. This proposed DEMO model showcases the potential to elevate the predictive performance in both short-term and long-term horizons, by providing dynamics insights into advanced learning algorithms. Evaluations on benchmark datasets such as NGSIM, MoCAD, HighD, and nuScenes show that DEMO outperforms SOTA models in both accuracy and efficiency, proving suitable for real-time applications. DEMO’s balanced approach to short-term and long-term prediction offers a promising path forward in enhancing AV safety and reliability.

\section*{Acknowledgement}
This research is supported by Science and Technology Development Fund of Macau SAR (File no. 0021/2022/ITP, 0081/2022/A2, 001/2024/SKL), Shenzhen-Hong Kong-Macau Science and Technology Program Category C (SGDX20230821095159012), State Key Lab of Intelligent Transportation System (2024-B001), Jiangsu Provincial Science and Technology Program (BZ2024055), and University of Macau (SRG2023-00037-IOTSC,
MYRG-GRG2024-00284-IOTSC).

\printcredits

% \bibliographystyle{cas-model2-names}
\bibliographystyle{model1-num-names}
\bibliography{cas-refs}


\end{sloppypar}
\end{document}